\documentclass{article}
\usepackage{ijcai22}

\usepackage{times}
\usepackage{soul}
\usepackage{url}
\usepackage[hidelinks]{hyperref}
\usepackage[utf8]{inputenc}
\usepackage[small]{caption}
\usepackage{graphicx}
\usepackage{amsmath}
\usepackage{amsthm}
\usepackage{booktabs}
\usepackage{algorithm}
\usepackage{algorithmic}
\usepackage{multirow}
\usepackage{subfigure}
\usepackage{amsmath}
\usepackage{diagbox}
\usepackage{color}
\graphicspath{{figure/}}

\title{SNCSE: Contrastive Learning for Unsupervised Sentence Embedding \\
       with Soft Negative Samples}

\author{Hao Wang$^{\dag,\ddag}$, Yangguang Li$^{\ddag}$, Zhen Huang$^{\dag}$, Yong Dou$^{\dag}$, Lingpeng Kong$^{\S}$, Jing Shao$^{\ddag}$
\affiliations{$^{\dag}$ National University of Defense Technology \\
              $^{\ddag}$ SenseTime\\
              $^{\S}$ The University of Hong Kong}
\emails{\{wanghao2, liyangguang\}@sensetime.com}
}

\begin{document}

\maketitle

\begin{abstract}
Unsupervised sentence embedding aims to obtain the most appropriate embedding for a sentence to reflect its semantic.
Contrastive learning has been attracting developing attention.
For a sentence, current models utilize diverse data augmentation  methods to generate positive samples, while consider other independent sentences as negative samples.
Then they adopt InfoNCE loss to pull the embeddings of positive pairs gathered, and push those of negative pairs scattered.
Although these models have made great progress on sentence embedding, we argue that they may suffer from \textit{feature suppression}.
The models fail to distinguish and decouple textual similarity and  semantic similarity.
And they may overestimate the semantic similarity of any pairs with similar textual regardless of the actual semantic difference between them.
This is because positive pairs in unsupervised contrastive learning come with similar and even the same textual through data augmentation.
To alleviate feature suppression, we propose contrastive learning for unsupervised sentence embedding with soft negative samples (SNCSE).
Soft negative samples share highly similar textual but have surely and apparently different semantic with the original samples.
Specifically, we take the negation of original sentences as soft negative samples, and propose Bidirectional Margin Loss (BML) to introduce them into traditional contrastive learning framework, which merely involves positive and negative samples.
Our experimental results show that SNCSE can obtain state-of-the-art performance on semantic textual similarity (STS) task with average Spearman's correlation coefficient of 78.97\% on BERT$_{base}$ and 79.23\% on RoBERTa$_{base}$. 
Besides, we adopt rank-based error analysis method to detect the weakness of SNCSE for future study.

\end{abstract}

\section{Introduction}

Pretrained language models, e.g.\;BERT \cite{devlin:bert} and RoBERTa \cite{yinhan:roberta}, have shown excellent performance on several natural language processing tasks with downstream fine-tuning.
However, it seems that they may fail to product appropriate sentence embedding to reflect the semantic \cite{reimers:sentence-bert,bohan:bert-flow,yan:consert}.
Recently, contrastive learning has been attracting developing attention for unsupervised sentence embedding \cite{yan:consert,gao:simcse,wu:esimcse}.
Contrastive learning firstly comes from computer vision field \cite{he:moco,chen:simclr,chen:mocov2}.
The main idea is to utilize diverse data augmentation  methods to generate positive pairs, while take two independent samples as negative pairs, and then adopt InfoNCE loss \cite{oord:infocse} to pull the embeddings of positive pairs gathered, and push those of negative pairs scattered.

\begin{figure}[!h]
\centerline{\subfigure[Spatial positions of different samples.]{\includegraphics[scale=0.55]{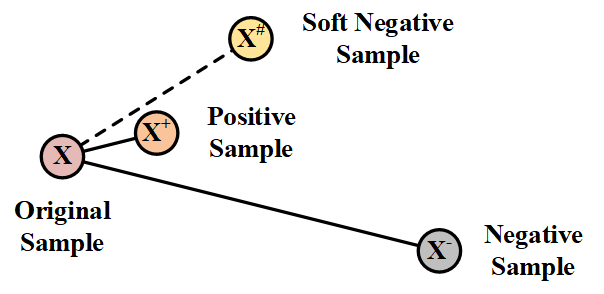}}}
\centerline{\subfigure[Textual sentences of different samples.]{\includegraphics[scale=0.5]{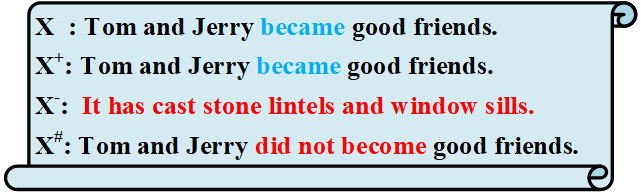}}}
\caption{Spatial positions and textual sentences of different samples.Previous unsupervised models aim to pull positive pairs gathered, and push negative pairs scattered. The positive sample may share the same textual with the original one with different embedding through cutoff \protect \cite{yan:consert} or dropout \protect \cite{gao:simcse}, while the negative one is totally another independent sentence. SNCSE tries to take the negation of original sentences as soft negative samples, and introduces them into traditional contrastive learning framework through BML loss.}
\label{motivation}
\end{figure}

\begin{figure*}
\centering
\centerline{\includegraphics[scale=0.5]{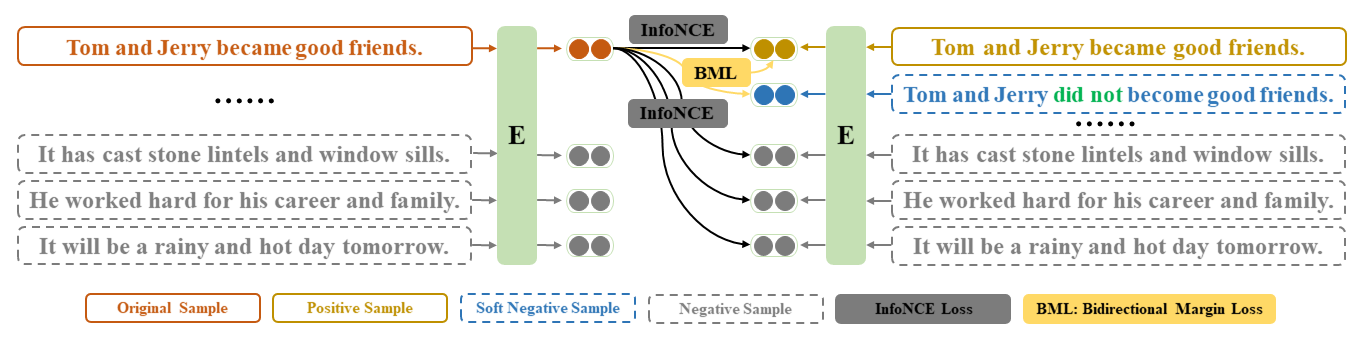}}
\caption{SNCSE framework. InfoNCE loss is utilized to distinguish positive samples from negative samples, while bidirectional margin loss is utilized to distinguish positive samples from soft negative samples. SNCSE does not contrast soft negative samples and negative samples due to the unsure relationship between these two.}
\label{framework}
\end{figure*}

Unsupervised sentence embedding takes similar strategy. 
As shown in Figure \ref{motivation}, positive pairs may share the same textual but have different embeddings through cutoff \cite{yan:consert} or dropout \cite{gao:simcse},  while negative pairs will come from two independent sentences which share less textual.
In this condition, we argue that this will lead to \textbf{feature suppression}, which has been discussed deeply in vision field \cite{joshua:feature-suppression1,ting:feature-suppression2}.
That is, the models fail to distinguish and decouple textual similarity and semantic similarity.
As a result, they may overestimate the semantic similarity of any pairs with similar textual regardless of the actual semantic difference between them.
And the models may underestimate the semantic similarity of pairs with less words in common. (Please refer to Section \ref{Error analysis} for several instances and detailed analysis.)
To \textbf{alleviate} feature suppression, we take following issues into consideration:

$\bullet$ whether there exist some \textbf{easily available} samples, which have highly similar textual but \textbf{surely and apparently} different semantic with the original samples

$\bullet$ how to introduce these samples into the traditional contrastive learning framework to enhance unsupervised sentence embedding, where the traditional framework only involves pure positive and negative samples

We call these samples \textbf{soft negative sample}. Here \textit{negative} means they have different semantics with the original samples, while \textit{soft} means they have high textual similarity with the original samples and cannot be merely considered as pure negative samples like other independent sentences.
Moreover, we propose contrastive learning for unsupervised sentence embedding with soft negative samples (SNCSE). 
The framework takes the negation of original sentences as soft negative samples, and proposes bidirectional margin loss (BML) to model the semantic difference between positive samples and soft negative samples, which both have highly similar textual with the original samples.
In this way, SNCSE can learn to distinguish and decouple textual similarity and  semantic similarity, and thus alleviate feature suppression.

We conduct our experiments on semantic textual similarity (STS) task. 
Our experimental results show that our SNCSE framework can achieve state-of-the-art performance on the task with diverse pretrained encoders, where average Spearman's correlation coefficient are respectively 78.97\% for BERT$_{base}$, 80.19\% for BERT$_{large}$, 79.23\% for RoBERTa$_{base}$ and 81.77\% for RoBERTa$_{large}$.
Moreover, our ablation study results indicate that SNCSE has learned to distinguish a sentence and its negation for the alleviation of feature suppression.
Besides, we adopt rank-based error analysis method to detect the weakness of SNCSE for future study.
Our code and models\footnote{\url{https://github.com/Sense-GVT/SNCSE}} have been released for the development of the field.

\section{SNCSE Framework} 

As shown in Figure \ref{framework}, in SNCSE framework, InfoNCE loss is utilized to distinguish positive samples from negative samples, while bidirectional margin loss is utilized to distinguish positive samples from soft negative samples.
SNCSE does not contrast soft negative samples and negative samples, since it is difficult to clarify the relationship between them.

In SNCSE, we take the negation of the original sentences as soft negative samples.
Specifically, the negated sentences are generated through a rule-based method.
Specifically, we utilize Spacy tools\footnote{https://spacy.io/} to parse the sentence to obtain the dependency syntax tree of the sentence, and the pos tag as well as the stem of each token. 
Then according to this information, we convert the sentences to its negation with correct syntax and clear semantic.
We mainly focus on explicit negation with negation words, which converts the original sample to \textit{``Tom and Jerry did not become good friends"}.
However, we do not consider implicit negation based on antonym, which may convert the original sample to \textit{``Tom and Jerry became bad friends"}.

Formally, in SNCSE, for a sentence, let \textit{X}, \textit{$X^+$}, \textit{$X^\#$} respectively denote the original sample, positive sample and soft negative sample.
Here, \textit{X} and \textit{$X^+$} share the same textual.
Inspired by recent studies on prompt learning \cite{ding:prompt1,yao:prompt2,liu:prompt3}, the three samples are encoded with pretrained language models, e.g.\;BERT and RoBERTa, within different prompts\footnote{We are also inspired by PromptBERT, which has been under review but withdrawn on OpenReview.net.}:
\begin{equation}
\centering
\begin{split}
The\;sentence\;:\;"\;X\;"\;means\;[MASK]. \\ 
The\;sentence\;of\;"\;X^+\;"\;means\;[MASK]. \\
The\;sentence\;of\;"\;X^\#\;"\;means\;[MASK]. \\
\end{split}
\end{equation}

We take the hidden state of the special [MASK] token \textit{h$_{[MASK]}$} as the embedding \textit{h} of the corresponding sample.
During training, like SimCSE \cite{gao:simcse}, we add an MLP layer with nonlinear activation function on \textit{h}$_{[MASK]}$ to obtain \textit{h}:
\begin{equation}
h = Tanh(MLP(h_{[MASK]}))
\end{equation}

InfoNCE loss is utilized to distinguish positive samples from negative samples:
\begin{equation}
\mathcal{L}_{InfoNCE} = -log \frac {e^{cos\_sim(h_i, h_i^{+})/\tau}}{\sum\limits_{j=1}^{N} e^{cos\_sim(h_i, h_j^+)/\tau}}
\end{equation}
where \textit{cos${\_}$sim} means cosine similarity, $\tau$ means temperature factor, while \textit{N} means batch size.

Then we calculate the cosine similarity  difference \textit{$\Delta$} between positive pairs and soft negative pairs:
\begin{equation}
\Delta = cos\_sim(h_i, h_i^\#) - cos\_sim(h_i,h_i^+) \\
\end{equation}
and  propose bidirectional margin loss (BML) to model the semantic similarity difference:
\begin{equation}
\mathcal{L}_{BML}= ReLU(\Delta + \alpha) + ReLU(-\Delta - \beta)
\end{equation}
BML loss aims to constraint $\Delta$ within this interval:
\begin{equation}
\Delta \in [-\beta, -\alpha]
\end{equation}
where $\alpha$ and $\beta$ respectively means the bottom and upper semantic similarity difference between positive pairs and soft negative pairs.

Finally, the total loss is the weighted summary of InfoNCE loss and bidirectional margin loss:
\begin{equation}
\mathcal{L}_{SNCSE} = \mathcal{L}_{InfoNCE} + \lambda \mathcal{L}_{BML}
\end{equation}
where the weight $\lambda$ reflects the importance of soft negative samples, and also adjusts the form difference of these two loss functions.

For comparison, we also consider to ignore soft negative samples, and merely take the negations as pure positive samples, or pure negative samples.
The corresponding positive loss \textit{PL} and negative loss \textit{NL} are respectively:
\begin{equation}
\begin{split}
\mathcal{L}_{PL} = -log \frac {e^{cos\_sim(h_i, h_i^{+})/\tau}}{\sum\limits_{j=1}^{N} e^{cos\_sim(h_i, h_j^+)/\tau}} -log \frac {e^{cos\_sim(h_i, h_i^{\#})/\tau}}{\sum\limits_{j=1}^{N} e^{cos\_sim(h_i, h_j^\#)/\tau}}
\end{split}
\label{positive loss}
\end{equation}
\begin{equation}
\mathcal{L}_{NL} = -log \frac {e^{cos\_sim(h_i, h_i^{+})/\tau}}{e^{cos\_sim(h_i, h_i^\#)/\tau} + \sum\limits_{j=1}^{N} e^{cos\_sim(h_i, h_j^+)/\tau} }
\label{negative loss}
\end{equation}

\section{Experiments}

\subsection{Evaluation Metric}

Following previous works \cite{yan:consert,gao:simcse,wu:esimcse}, we evaluate SNCSE on semantic textual similarity (STS) task.
STS task aims to measure the semantic similarity of sentence pairs, and is one of the most widely used  benchmark datasets for evaluation of unsupervised sentence embedding.
The task include 7 subtasks in total, which are respectively STS 12-16, STS-B, and SICK-R \cite{agirre:sts12,agirre:sts13,agirre:sts14,agirre:sts15,agirre:sts16,cer:stsb,marco:sickr}.
Each subtask contains several sentence pairs, and the semantic similarity of a sentence pair is labeled as a number between 0.0 and 5.0, where higher score means higher semantic similarity.
The detailed information of these subtasks are summarized in Table \ref{staticitc information}.
During evaluation, we calculate cosine similarity of each pair, and focus on the test sets.
For a pair, golden semantic similarity ranges from 0.0 to 5.0, while cosine similarity ranges from -1.0 to 1.0.
To deal with the range difference, rank-based Spearman's correlation coefficient is adopted.

\begin{table}[!h]
\centering
\begin{tabular}{ccccc}
\hline
\textbf{Task} & \textbf{STS12} & \textbf{STS13} & \textbf{STS14}  & \textbf{STS15} \\ \hline
train         & 0              & 0              & 0               & 0              \\
dev           & 0              & 0              & 0               & 0              \\
test          & 3108           & 1500           & 3750            & 3000           \\ \hline
\textbf{Task} & \textbf{STS16} & \textbf{STS-B} & \textbf{SICK-R} & \textbf{}      \\ \hline
train         & 0              & 5749           & 4500            &                \\
dev           & 0              & 1500           & 500             &                \\
test          & 1186           & 1379           & 4927            &                \\ \hline
\end{tabular}
\caption{Statistic information of STS task.}
\label{staticitc information}
\end{table}

Specifically, for a subtask with \textit{n} sentence pairs, we first sort the golden similarity to obtain the golden rank R$_G$ for each sentence pair.
And we also calculate and sort estimated cosine similarity to obtain estimated rank R$_E$ for each pairs.
Then, rank-based Spearman's correlation coefficient $\rho$ for the subtask is defined as:
\begin{equation}
\rho = 1 - \frac {\sum\limits_{i=1}^{n} {(R_E^i-R_G^i)}^2} {\frac {1} {6} (n^3-n)}
\end{equation}
Here $\rho$ ranges from -1.0 to 1.0, where higher value means more advanced sentence embedding and semantic comprehension ability of the model.
We calculate Spearman's correlation coefficient $\rho$ for each subtask independently and take the average of $\rho$ on all subtasks as main evaluation metric.

Moreover, we conduct rank-based error analysis to detect the weakness of our SNCSE.
For each sentence pair, we roughly\footnote{Here the rank difference of a pair is related to not only the pair itself, but also other pairs of the same subtask.} take the rank difference between $R_G$ and $R_E$:
\begin{equation}
(R_E^i-R_G^i)^2
\label{error}
\end{equation}
as the semantic estimation error of the pair.
Then we focus on sentences pairs with the most severe semantic estimation error for further analysis. 

\begin{table*}[h]
\centering
\begin{tabular}{ccccccccc}
\hline
\textbf{Model}       & \textbf{STS12} & \textbf{STS13} & \textbf{STS14} & \textbf{STS15} & \textbf{STS16} & \textbf{STS-B} & \textbf{SICK-R} & \textbf{Avg.}  \\ \hline
ConSERT-BERT$_{base}\heartsuit$     & 64.64          & 78.49          & 69.07          & 79.72          & 75.95          & 73.97          & 67.31           & 72.74          \\
SimCSE-BERT$_{base}\clubsuit$      & 68.40          & 82.41          & 74.38          & 80.91          & 78.56          & 76.85          & 72.23           & 76.25          \\
ESimCSE-BERT$_{base}\spadesuit$      & \textbf{73.40} & 83.27          & \textbf{77.25} & 82.66          & 78.81          & 80.17          & 72.30           & 78.27          \\ \hline 
SNCSE-BERT$_{base}$      & 70.67          & \textbf{84.79} & 76.99          & 83.69          & \textbf{80.51} & \textbf{81.35} & \textbf{74.77}  & \textbf{78.97} \\
w.o. BML Loss            & 71.58          & 84.68          & 76.96          & \textbf{83.78} & 79.40          & 80.46          & 70.64           & 78.21          \\ \hline \hline
ConSERT-BERT$_{large}\heartsuit$    & 70.69          & 82.96          & 74.13          & 82.78          & 76.66          & 77.53          & 70.37           & 76.45          \\
SimCSE-BERT$_{large}\clubsuit$     & 70.88          & 84.16          & 76.43          & 84.50          & 79.76          & 79.26          & 73.88           & 78.41          \\
ESimCSE-BERT$_{large}\spadesuit$    & \textbf{73.21} & 85.37          & 77.73          & 84.30          & 78.92          & 80.73          & 74.89           & 79.31          \\ \hline 
SNCSE-BERT$_{large}$     & 71.94          & 86.66          & \textbf{78.84} & \textbf{85.74} & \textbf{80.72} & \textbf{82.29} & \textbf{75.11}  & \textbf{80.19} \\ 
w.o. BML Loss            & 73.18          & \textbf{86.74} & 78.44          & 85.66          & 79.69          & 81.98          & 71.69           & 79.62          \\ \hline \hline
SimCSE-RoBERTa$_{base}\clubsuit$   & 70.16          & 81.77          & 73.24          & 81.36          & 80.65          & 80.22          & 68.56           & 76.57          \\
ESimCSE-RoBERTa$_{base}\spadesuit$  & 69.90          & 82.50          & 74.68          & 83.19          & 80.30          & 80.99          & 70.54           & 77.44          \\ \hline 
SNCSE-RoBERTa$_{base}$    & 70.62          & \textbf{84.42} & \textbf{77.24} & \textbf{84.85} & \textbf{81.49} & \textbf{83.07} & \textbf{72.92}  & \textbf{79.23} \\
w.o. BML Loss            & \textbf{71.22} & 84.10          & 76.44          & 84.35          & 80.80          & 81.77          & 69.82           & 78.36          \\ \hline \hline
SimCSE-RoBERTa$_{large}\clubsuit$  & 72.86          & 83.99          & 75.62          & 84.77          & 81.80          & 81.98          & 71.26           & 78.90          \\
ESimCSE-RoBERTa$_{large}\spadesuit$ & 73.20          & 84.93          & 76.88          & 84.86          & 81.21          & 82.79          & 72.27           & 79.45          \\ \hline 
SNCSE-RoBERTa$_{large}$   & 73.71          & \textbf{86.73} & \textbf{80.35} & \textbf{86.80} & \textbf{83.06} & \textbf{84.31} & \textbf{77.43}  & \textbf{81.77} \\
w.o. BML Loss            & \textbf{73.73} & 85.66          & 79.00          & 86.34          & 81.78          & 83.64          & 72.10           & 80.32          \\ \hline
\end{tabular}
\caption{Sentence embedding performance on  semantic textual similarity (STS) test sets, in terms of Spearman's correlation coefficient. $\heartsuit$ denotes the results from Yan \textit{et al.} \protect \shortcite{yan:consert}, $\clubsuit$ denotes the results from Gao \textit{et al.} \protect \shortcite{gao:simcse}, while $\spadesuit$ denotes the results from Wu \textit{et al.} \protect \shortcite{wu:esimcse}. We also conduct ablation study to verify the influence of BML loss, and \textit{w.o.} means \textit{without}.}
\label{main results}
\end{table*}

\subsection{Experiment Setup}

Our framework is implemented through Pytorch 1.8.0 and hugging face transformers\footnote{https://huggingface.co}.
All our experiments are conducted on a computation node with Nvidia A100 GPUs.
For all experiments, we adopt AdamW optimizer.
For all experiments, we keep a random number seed of 42, and set dropout rate as 0.1, temperature factor $\tau$ as 0.05, $\alpha$ as 0.1 and $\beta$ as 0.3.
Besides, we set $\lambda$ as 1e-3 for BERT encoder and 5e-4 for RoBERTa encoder.
For base models, we adopt batch size of 256 and initial learning rate of 1e-5, while for large models, we adopt batch size of 128 and learning rate of 5e-6 to avoid out of memory problem.
Following SimCSE \cite{gao:simcse} and ESimCSE \cite{wu:esimcse}, we adopt 1 million sentences\footnote{\url{https://huggingface.co/datasets/princeton-nlp/datasets-for-simcse/blob/main/wiki1m_for_simcse.txt}} randomly drawn from English Wikipedia as train sentences.
And we also train 1 epoch, evaluate every 125 steps and choose model parameters with the highest performance on STSB development set.

\subsection{Main Results}

Our main experimental results are shown in Table \ref{main results}.
Our SNCSE framework obtains the best performance with the highest average Spearman's correlation coefficient on all encoders, which are respectively 78.97\% for BERT$_{base}$, 80.19\% for BERT$_{large}$, 79.23\% for RoBERTa$_{base}$ and 81.77\% for RoBERTa$_{large}$.
Compared with ESimCSE \cite{wu:esimcse}, which is previous state-of-the-art model, SNCSE respectively obtain 0.7\%, 0.87\%, 1.79\% and 2.32\% absolute increase on average Spearman's correlation coefficient.
Moreover, we also conduct ablation study to detect the influence of BML loss.
For the four encoders, with BML loss, our SNCSE respectively obtain 0.76\%, 0.57\%, 0.87\% and 1.45\% absolute increase on average Spearman's correlation coefficient.
These indicate the strength of SNCSE and BML loss.

\section{Ablation Study}

\subsection{Role of the Negation}

For comparison, we try to change the role of the negation with different loss functions.
The experimental results with BERT$_{base}$ encoder are shown in Table \ref{role of negation}.
Traditional InfoNCE loss merely takes positive and negative pairs into consideration and ignore the negation, and obtains average Spearman's correlation coefficient of 78.21\%.
Positive loss in Equation \ref{positive loss} takes the negation as pure positive samples, and obtains 77.11\%, while negative loss in Equation \ref{negative loss} takes the negation as pure negative samples and obtains 74.37\%.
SNCSE takes the negation as soft negative samples, contrasts them with the positive samples through BML loss, and obtains the highest Spearman's correlation coefficient of 78.97\%.
Therefore, it can be concluded that SNCSE considers the negation in a more appropriate way.

\begin{table}[!h]
\centering
\begin{tabular}{ccccc}
\hline
\textbf{Loss} & PL    & NL    & InfoNCE & SNCSE          \\ \hline
\textbf{Avg.} & 77.11 & 74.37 & 78.21   & \textbf{78.97} \\ \hline
\end{tabular}
\caption{Influence of the role of the negation. \textit{PL} takes the negation as pure positive samples, \textit{NL} take as pure negative sample, \textit{InfoNCE} ignores the negation, while \textit{SNCSE} takes as soft negative samples.}
\label{role of negation}
\end{table}

\begin{figure*}[!h]
\subfigure[$\alpha=0.0$]{
\begin{minipage}[t]{0.35\linewidth}
\centerline{\includegraphics[scale=0.455]{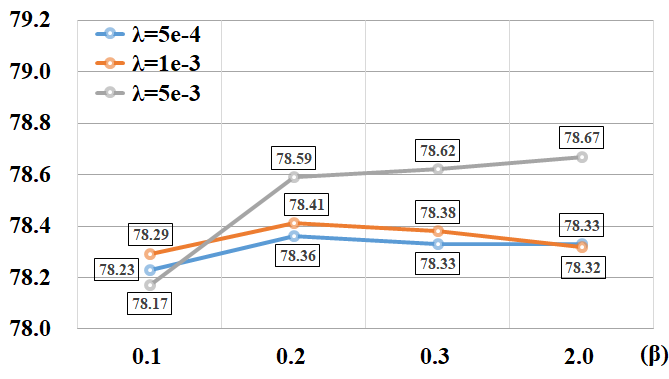}}
\end{minipage}
}
\subfigure[$\alpha=0.1$]{
\begin{minipage}[t]{0.35\linewidth}
\centerline{\includegraphics[scale=0.455]{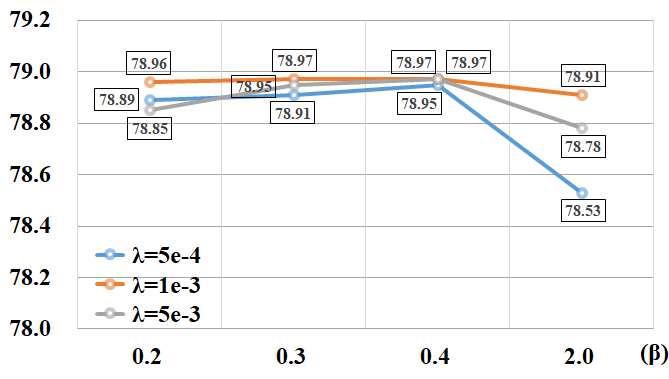}}
\end{minipage}
}
\subfigure[$\alpha=0.2$]{
\begin{minipage}[t]{0.285\linewidth}
\centerline{\includegraphics[scale=0.455]{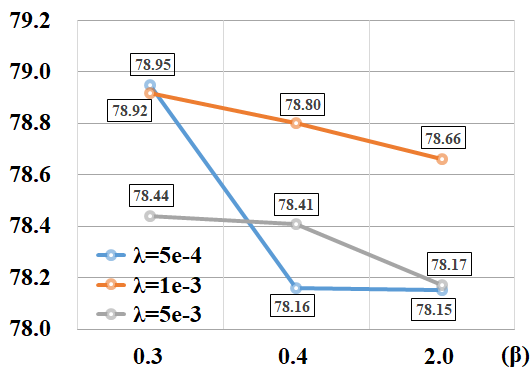}}
\end{minipage}
}
\caption{Influence of hyperparameters of BML loss. $\lambda$ denotes the importance of soft negative sample, and adjusts the form difference between InfoNCE loss and margin loss. $\alpha$ and $\beta$ denotes the bottom and upper difference between positive pairs and soft negative pairs. The upper difference is ignored with $\beta$ as 2.0.} 
\label{hyperparameters}
\end{figure*}

\subsection{Influence of Hyperparameters in BML Loss}

In SNCSE, we introduce BML loss with 3 main hyperparameters to constraint the semantic similarity difference between positive pairs and soft negative pairs.
$\lambda$ denotes the importance soft negative samples, and adjusts the form difference between InfoNCE loss and margin loss.
$\alpha$ denotes the bottom difference between positive pairs and soft negative pairs, and a higher value requires more apparent difference.
$\beta$ denotes the upper difference, and upper difference is ignored with $\beta$ as 2.0, since cosine similarity ranges from -1.0 to 1.0.

As shown in Figure \ref{hyperparameters}, it seems that $\alpha$ is of the most importance.
With different $\lambda$ and $\beta$, setting $\alpha$ as 0.1 shows the best performance.
This means $\alpha$ as 0.1 is more appropriate to reflect the difference between positive pairs and soft negative pairs.
Besides, with $\alpha$ as 0.1, the performance gets slight increase with $\beta$ from 0.2 to 0.4, but gets apparent  decrease with $\beta$ as 2.0.
With $\alpha$ as 0.2, the performance also gets apparent decrease with the increase of $\beta$.
This may indicate it is better to set meaningful upper difference between positive pairs and soft negative pairs.
Moreover, $\lambda$ also shows great influence on the performance.

\subsection{Alleviation of Feature Suppression}

\begin{figure}[!h]
\centerline{\subfigure[Cosine similarity distribution of \textit{soft negative} pairs.]{\includegraphics[scale=0.55]{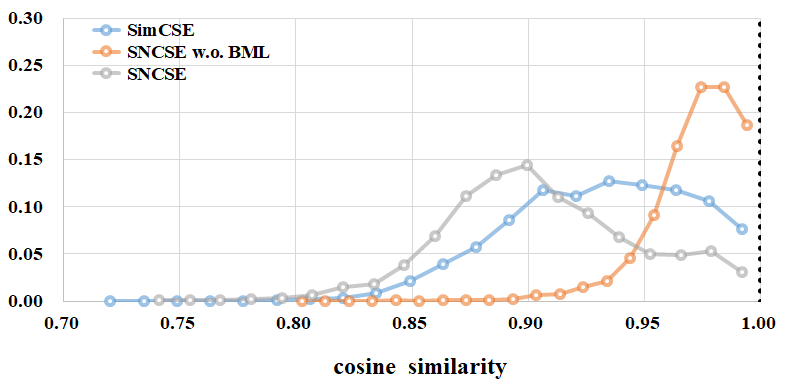}}}
\centerline{\subfigure[Cosine similarity distribution of \textit{negative} pairs.]{\includegraphics[scale=0.55]{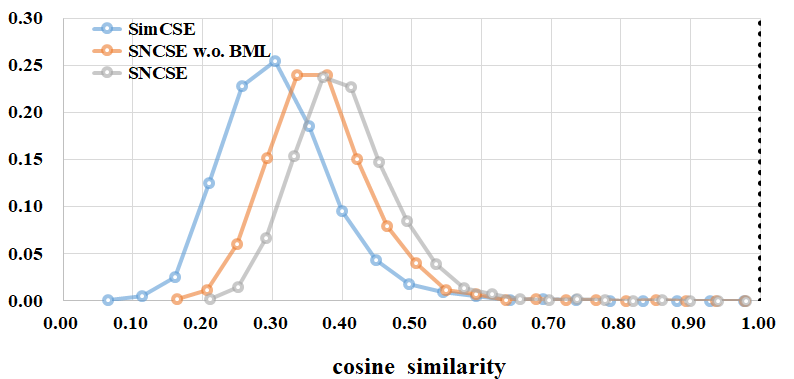}}}
\caption{Cosine similarity distribution on different pairs. The vertical black point line means the default similarity of positive pairs is 1.0. We take soft negative samples for instance to detect the influence of SNCSE on the alleviation of feature suppression.}
\label{distribution of cosine similarity}
\end{figure}

To further detect the influence of SNCSE on the alleviation of feature suppression, we take soft negative samples for instance. 
Specifically, we randomly select 5000 sentences from STS task, and visualize the distribution of cosine similarity of different models on positive pairs, negative pairs, and soft negative pairs.
For each sentence, we take itself as positive pair, and set default cosine similarity as 1.
Then we take the negation of the sentence to construct soft negative pair, and take another randomly selected sentence as negative pair\footnote{We do not take the original sentence pairs in STS task, since several pairs have high semantic similarity and should not be considered as negative pairs.}.
We also take  SimCSE \cite{gao:simcse} for comparison.

\begin{table*}[h]
\centering
\begin{tabular}{ccccc}
\hline
\textbf{ID} & \textbf{Sentence1}                                                                                        & \textbf{Sentence2}                                                                                            & \textbf{R$_{G}$} & \textbf{R$_{E}$} \\ \hline
\multicolumn{5}{c}{STS14}                                                                                                                                                                                                                                                             \\ \hline
1           & \begin{tabular}[c]{@{}c@{}}I wood have asked her \\ if she is up for a \textcolor{red}{treesome}.\end{tabular}             & \begin{tabular}[c]{@{}c@{}}I wood have at least asked if \\ a \textcolor{red}{threesome} was out of the question.\end{tabular} & 3564              & 584                     \\
2           & \begin{tabular}[c]{@{}c@{}}``being against nukes" \textcolor{green}{does not mean}\\  ``not wanting to use nukes"\end{tabular} & \begin{tabular}[c]{@{}c@{}}``being against using nukes" \textcolor{green}{means} \\ ``not wanting to use nukes".\end{tabular}      & 799               & 3695                    \\
3           & I \textcolor{green}{oppose} the death penalty.                                                                               & This is why I \textcolor{green}{support} the death penalty .                                                                     & 636               & 3475                    \\
4           & \textcolor{red}{fundimental} difference, please?                                                                           & \textcolor{red}{fundamental} difference?                                                                                       & 3324              & 535                     \\
5           & what conviction \textcolor{blue}{would that be}?                                                                            & what conviction \textcolor{blue}{where you talking about before}?                                                               & 3164              & 495                     \\ \hline
\multicolumn{5}{c}{SICK-R}                                                                                                                                                                                                                                                            \\ \hline
1           & A man is \textcolor{blue}{doing pull-ups}                                                                                   & The man is \textcolor{blue}{exercising}                                                                                         & 4187              & 304                     \\
2           & A \textcolor{cyan}{person} is kicking a \textcolor{cyan}{monkey}                                                             & A \textcolor{cyan}{monkey} is kicking \textcolor{cyan}{a person}                                                                                 & 1140              & 4869                    \\
3           & \begin{tabular}[c]{@{}c@{}}A gloved \textcolor{cyan}{person} is kicking \\ at the hand of a \textcolor{cyan}{monkey}\end{tabular}             & \begin{tabular}[c]{@{}c@{}}A \textcolor{cyan}{monkey} is kicking at the \\ gloved hand of a \textcolor{cyan}{person}\end{tabular}                 & 1328              & 4859                    \\
4           & Ferrets are climbing \textcolor{green}{up} a shelf                                                                           & Ferrets are climbing \textcolor{green}{down from} a shelf                                                                        & 1129              & 4615                    \\
5           & A small \textcolor{cyan}{girl} is \textcolor{cyan}{riding in a toy car}                                                                       & A small \textcolor{cyan}{toy girl} is \textcolor{cyan}{in a riding car}                                                                          & 871               & 4354                    \\ \hline
\end{tabular}
\caption{Top 5 error pairs of STS14 and SICK-R. The similarity scores are sorted from low to high and the error is defined as Equation \ref{error}. For a sentence pair, the similarity score is overestimated with \textit{R$_{E}$} higher than \textit{R$_{G}$}, and vice versa. We mark possible key issue with different colors, where \textcolor{red}{red} involves typos, \textcolor{green}{green} involves negation logic, \textcolor{blue}{blue} involves textual independency, while \textcolor{cyan}{cyan} involves word order.}
\label{error analysis}
\end{table*}

The results are shown in Figure \ref{distribution of cosine similarity}.
For positive pairs, the default similarity is 1.0.
For negative pairs, the similarity peak of SimCSE, SNCSE without BML loss and SNCSE respectively appear at about 0.30, 0.38 and 0.37, which are far lower than 1.0.
This indicates that all models can clearly distinguish positive pairs from negative pairs.
For soft negative pairs, SimCSE, with similarity peak at 0.93, can somewhat distinguish soft negative pairs from positive pairs.
However, SNCSE without BML loss, with the peak at 0.98, can hardly distinguish these two, although it highly outperforms SimCSE.
It seems that the model simply takes soft negative pairs as pure positive pairs because of the high textual similarity between them, and can hardly recognize the semantic difference between them with severe feature suppression.
Nevertheless, SNCSE, with the peak at about 0.90, is able to clearly model the difference between these two , thus alleviate feature suppression and obtain better performance. 

\section{Discussion}
\label{Error analysis}

To further detect the weakness of SNCSE, we propose rank-based error analysis.
Specifically, we focus on top 5 error pairs, which have the highest rank  estimation error as defined in  Equation \ref{error}, on STS14 and SICK-R, since these two have the most sentence pairs.
As shown in Table \ref{error analysis}, the main errors can be concluded as 4 kinds:

$\bullet$ \textbf{Typos}. Like the No.1 and No.4 pairs of STS14, the pairs themselves are annotated as high similarity. However, SNCSE underestimates the semantic similarity, since the model cannot understand the relationship between the word \textit{``threesome"} and the typo \textit{``treesome"}, and the word \textit{``fundamental"} and the typo \textit{``fundimental"}. This can be considered as annotation error. 

$\bullet$ \textbf{Negation logic}. Like No.2, No.3 of STS14, and No.4 of SICK-R, one sentence of these pairs is the negation of another one. Differently, No.2 of STS14 involves explicit negation with the word ``not", while the other two involve implicit negation based on antonyms. It is because of feature suppression that SNCSE overestimates the similarity.

$\bullet$ \textbf{Textual Independency}. Like No.5 of STS14 and No.1 of STS, the pairs are of high semantic similarity. However, the pairs share less textual and look like totally independent sentences. Therefore, SNCSE, which takes two independent sentences as negative pairs, tends to underestimate the semantic similarity of these pairs due to feature suppression.

$\bullet$ \textbf{Word Order}. Like No.2, No.3 and No.5 of SICK-R, the pairs share the same words but come with different word order. Actually, they have great semantic difference. However, SNCSE ignores the semantic difference due to the extremely similar textual. This is also caused by feature suppression.

In summary, it is far to totally handle feature suppression, although SNCSE has made some progress.
The main problem is to distinguish and decouple textual similarity and  semantic similarity.
However, this will be extremely difficult for \textbf{self-attention based} transformer encoders, e.g.\;BERT and RoBERTa, where the embedding of a sentence comes from the weighted summary of the embedding of the words.
Especially, it is far more difficult to accurately model the relationship between word order and sentence semantic with the position embedding. 
For instance, “Tom and Jerry became good friends” and “Jerry and Tom became good friends” share similar semantic, while the pairs involving \textit{word order} have apparently different semantics.
Therefore, it is necessary to develop stronger models to deal with feature suppression.

\section{Related Work}

Reimers and Gurevych \shortcite{reimers:sentence-bert}, Li \textit{et al.} \shortcite{bohan:bert-flow}, and  Yan \textit{et al.} \shortcite{yan:consert} find that vanilla sentence embeddings from pretrained language models are of low quality to reflect the semantic, where the embeddings of any sentence pairs are of high similarity.
To handle the problem, Reimers and Gurevych \shortcite{reimers:sentence-bert} propose Sentence-BERT, which trains the encoders through textual entailment tasks in a supervised way.
Early unsupervised sentence embedding models include BERT-flow \cite{bohan:bert-flow} and BERT-Whitening \cite{su:bert-whitening}, which are respectively based on reversible transformation and whitening operation.
Yan \textit{et al.} \shortcite{yan:consert} fistly introduce contrastive learning for unsupervised sentence embedding, and propose several data augmentation methods.
Gao \textit{et al.} \shortcite{gao:simcse} propose SimCSE, which considers dropout operation as a kind of data augmentation method.
Wu \textit{et al.} \shortcite{wu:esimcse} point out that SimCSE has the bias to overestimate the semantic similarity of sentence pairs with the same length.
To eliminate the bias, they propose ESimCSE with word repetition as data augmentation.
They also introduce momentum contrastive \cite{he:moco,chen:mocov2} for sentence embedding.
Differently, in this paper, we mainly focus on feature suppression.

\section{Conclusion}
In this paper, we propose SNCSE with bidirectional margin loss (BML) to introduce soft negative samples to enhance sentence embedding.
Our model makes significant progress and obtain state-of-the-art performance on STS task with all encoders.
Our ablation study results show the importance of BML loss and the improvement on alleviating feature suppression.
We also propose rank-based error analysis method to detect the weakness of SNCSE.
In the future, we will try to study much stronger models to deal with feature suppression.

\bibliographystyle{named}
\bibliography{SNCSE}

\begin{thebibliography}{}

\bibitem[\protect\citeauthoryear{Agirre \bgroup \em et al.\egroup
  }{2012}]{agirre:sts12}
Eneko Agirre, Daniel~M. Cer, Mona~T. Diab, and Aitor~Gonzalez Agirre.
\newblock Semeval-2012 task 6: A pilot on semantic textual similarity.
\newblock In {\em SemEval 2012}, pages 385--393. ACL, 2012.

\bibitem[\protect\citeauthoryear{Agirre \bgroup \em et al.\egroup
  }{2013}]{agirre:sts13}
Eneko Agirre, Daniel~M. Cer, Mona~T. Diab, Aitor~Gonzalez Agirre, and Weiwei
  Guo.
\newblock *sem 2013 shared task: Semantic textual similarity.
\newblock In {\em *SEM 2013}, pages 32--43. ACL, 2013.

\bibitem[\protect\citeauthoryear{Agirre \bgroup \em et al.\egroup
  }{2014}]{agirre:sts14}
Eneko Agirre, Carmen Banea, Claire Cardie, Daniel~M. Cer, Mona~T. Diab,
  Aitor~Gonzalez Agirre, Weiwei Guo, Rada Mihalcea, German Rigau, and Janyce
  Wiebe.
\newblock Semeval-2014 task 10: Multilingual semantic textual similarity.
\newblock In {\em SemEval 2014}, pages 81--91. ACL, 2014.

\bibitem[\protect\citeauthoryear{Agirre \bgroup \em et al.\egroup
  }{2015}]{agirre:sts15}
Eneko Agirre, Carmen Banea, Claire Cardie, Daniel~M. Cer, Mona~T. Diab,
  Aitor~Gonzalez Agirre, Weiwei Guo, Inigo~Lopez Gazpio, Montse Maritxalar,
  Rada Mihalcea, German Rigau, Larraitz Uria, and Janyce Wiebe.
\newblock Semeval-2015 task 2: Semantic textual similarity, english, spanish
  and pilot on interpretability.
\newblock In {\em SemEval 2015}, pages 252--263. ACL, 2015.

\bibitem[\protect\citeauthoryear{Agirre \bgroup \em et al.\egroup
  }{2016}]{agirre:sts16}
Eneko Agirre, Carmen Banea, Daniel~M. Cer, Mona~T. Diab, Aitor
  Gonzalez{-}Agirre, Rada Mihalcea, German Rigau, and Janyce Wiebe.
\newblock Semeval-2016 task 1: Semantic textual similarity, monolingual and
  cross-lingual evaluation.
\newblock In {\em SemEval 2016}, pages 497--511. ACL, 2016.

\bibitem[\protect\citeauthoryear{Cer \bgroup \em et al.\egroup
  }{2017}]{cer:stsb}
Daniel~M. Cer, Mona~T. Diab, Eneko Agirre, Inigo Lopez{-}Gazpio, and Lucia
  Specia.
\newblock Semeval-2017 task 1: Semantic textual similarity multilingual and
  crosslingual focused evaluation.
\newblock In {\em SemEval 2017}, pages 1--14. ACL, 2017.

\bibitem[\protect\citeauthoryear{Chen \bgroup \em et al.\egroup
  }{2020a}]{chen:simclr}
Ting Chen, Simon Kornblith, Mohammad Norouzi, and Geoffrey~E. Hinton.
\newblock A simple framework for contrastive learning of visual
  representations.
\newblock In {\em ICML 2020}, pages 1597--1607. PMLR, 2020.

\bibitem[\protect\citeauthoryear{Chen \bgroup \em et al.\egroup
  }{2020b}]{chen:mocov2}
Xinlei Chen, Haoqi Fan, Ross~B. Girshick, and Kaiming He.
\newblock Improved baselines with momentum contrastive learning.
\newblock {\em CoRR}, abs/2003.04297, 2020.

\bibitem[\protect\citeauthoryear{Chen \bgroup \em et al.\egroup
  }{2021}]{ting:feature-suppression2}
Ting Chen, Calvin Luo, and Lala Li.
\newblock Intriguing properties of contrastive losses.
\newblock In {\em NIPS 2021}. NeurIPS, 2021.

\bibitem[\protect\citeauthoryear{Devlin \bgroup \em et al.\egroup
  }{2019}]{devlin:bert}
Jacob Devlin, Ming~Wei Chang, Kenton Lee, and Kristina Toutanova.
\newblock Bert: Pre-training of deep bidirectional transformers for language
  understanding.
\newblock In {\em NAACL 2019}, pages 4171--4186. ACL, 2019.

\bibitem[\protect\citeauthoryear{Ding \bgroup \em et al.\egroup
  }{2021}]{ding:prompt1}
Ning Ding, Yulin Chen, Xu~Han, Guangwei Xu, Pengjun Xie, Hai~Tao Zheng, Zhiyuan
  Liu, Juanzi Li, and Hong~Gee Kim.
\newblock Prompt-learning for fine-grained entity typing.
\newblock {\em CoRR}, abs/2108.10604, 2021.

\bibitem[\protect\citeauthoryear{Gao \bgroup \em et al.\egroup
  }{2021}]{gao:simcse}
Tianyu Gao, Xingcheng Yao, and Danqi Chen.
\newblock Simcse: Simple contrastive learning of sentence embeddings.
\newblock In {\em EMNLP 2021}, pages 6894--6910. ACL, 2021.

\bibitem[\protect\citeauthoryear{He \bgroup \em et al.\egroup }{2020}]{he:moco}
Kaiming He, Haoqi Fan, Yuxin Wu, Saining Xie, and Ross~B. Girshick.
\newblock Momentum contrast for unsupervised visual representation learning.
\newblock In {\em CVPR 2020}, pages 9726--9735. IEEE/CVF, 2020.

\bibitem[\protect\citeauthoryear{Li \bgroup \em et al.\egroup
  }{2020}]{bohan:bert-flow}
Bohan Li, Hao Zhou, Junxian He, Mingxuan Wang, Yiming Yang, and Lei Li.
\newblock On the sentence embeddings from pre-trained language models.
\newblock In {\em EMNLP 2020}, pages 9119--9130. ACL, 2020.

\bibitem[\protect\citeauthoryear{Liu \bgroup \em et al.\egroup
  }{2019}]{yinhan:roberta}
Yinhan Liu, Myle Ott, Naman Goyal, Jingfei Du, Mandar Joshi, Danqi Chen, Omer
  Levy, Mike Lewis, Luke Zettlemoyer, and Veselin Stoyanov.
\newblock Roberta: {A} robustly optimized {BERT} pretraining approach.
\newblock {\em CoRR}, abs/1907.11692, 2019.

\bibitem[\protect\citeauthoryear{Liu \bgroup \em et al.\egroup
  }{2021}]{liu:prompt3}
Pengfei Liu, Weizhe Yuan, Jinlan Fu, Zhengbao Jiang, Hiroaki Hayashi, and
  Graham Neubig.
\newblock Pre-train, prompt, and predict: A systematic survey of prompting
  methods in natural language processing.
\newblock {\em CoRR}, abs/2107.13586, 2021.

\bibitem[\protect\citeauthoryear{Marelli \bgroup \em et al.\egroup
  }{2014}]{marco:sickr}
Marco Marelli, Stefano Menini, Marco Baroni, Luisa Bentivogli, Raffaella
  Bernardi, and Roberto Zamparelli.
\newblock A sick cure for the evaluation of compositional distributional
  semantic models.
\newblock In {\em LREC 2014}, pages 216--223. ELRA, 2014.

\bibitem[\protect\citeauthoryear{Oord \bgroup \em et al.\egroup
  }{2018}]{oord:infocse}
Aaron Van~Den Oord, Yazhe Li, and Oriol Vinyals.
\newblock Representation learning with contrastive predictive coding.
\newblock {\em CoRR}, abs/1807.03748, 2018.

\bibitem[\protect\citeauthoryear{Reimers and
  Gurevych}{2019}]{reimers:sentence-bert}
Nils Reimers and Iryna Gurevych.
\newblock Sentence-bert: Sentence embeddings using siamese bert-networks.
\newblock In {\em EMNLP 2019}, pages 3980--3990. ACL, 2019.

\bibitem[\protect\citeauthoryear{Robinson \bgroup \em et al.\egroup
  }{2021}]{joshua:feature-suppression1}
Joshua Robinson, Li~Sun, Ke~Yu, Kayhan Batmanghelich, Stefanie Jegelka, and
  Suvrit Sra.
\newblock Can contrastive learning avoid shortcut solutions?
\newblock In {\em NIPS 2021}. NeurIPS, 2021.

\bibitem[\protect\citeauthoryear{Su \bgroup \em et al.\egroup
  }{2021}]{su:bert-whitening}
Jianlin Su, Jiarun Cao, Weijie Liu, and Yangyiwen Ou.
\newblock Whitening sentence representations for better semantics and faster
  retrieval.
\newblock {\em CoRR}, abs/2103.15316, 2021.

\bibitem[\protect\citeauthoryear{Wu \bgroup \em et al.\egroup
  }{2021}]{wu:esimcse}
Xing Wu, Chaochen Gao, Liangjun Zang, Jizhong Han, Zhongyuan Wang, and Songlin
  Hu.
\newblock Esimcse: Enhanced sample building method for contrastive learning of
  unsupervised sentence embedding.
\newblock {\em CoRR}, abs/2109.04380, 2021.

\bibitem[\protect\citeauthoryear{Yan \bgroup \em et al.\egroup
  }{2021}]{yan:consert}
Yuanmeng Yan, Rumei Li, Sirui Wang, Fuzheng Zhang, Wei Wu, and Weiran Xu.
\newblock Consert: A contrastive framework for self-supervised sentence
  representation transfer.
\newblock In {\em ACL 2021}, pages 5065--5075. ACL, 2021.

\bibitem[\protect\citeauthoryear{Yao \bgroup \em et al.\egroup
  }{2021}]{yao:prompt2}
Yuan Yao, Ao~Zhang, Zhengyan Zhang, Zhiyuan Liu, Tat~Seng Chua, and Maosong
  Sun.
\newblock Cpt: Colorful prompt tuning for pre-trained vision-language models.
\newblock {\em CoRR}, abs/2109.11797, 2021.

\end{thebibliography}
\end{document}